\documentclass{article}

\usepackage[final,main]{neurips_2026}

\usepackage[utf8]{inputenc} %
\usepackage[T1]{fontenc}    %
\usepackage{hyperref}       %
\usepackage{url}            %
\usepackage{booktabs}       %
\usepackage{amsfonts}       %
\usepackage{nicefrac}       %
\usepackage{microtype}      %
\usepackage{xcolor}         %
\usepackage{amsmath}        %
\usepackage{amssymb}        %
\usepackage{algorithm}      %
\usepackage{algorithmic}    %
\usepackage{multirow}       %
\usepackage{graphicx}       %
\usepackage{placeins}       %
\usepackage{tikz}           %
\usepackage{pgfplots}       %
\pgfplotsset{compat=1.18}
\usepgfplotslibrary{groupplots}
\usetikzlibrary{positioning, shapes.geometric, shapes.misc}
\graphicspath{{./}}

\usepackage{titlesec}
\titlespacing*{\section}{0pt}{*0.6}{*0.5}
\titlespacing*{\subsection}{0pt}{*0.8}{*0.4}
\titlespacing*{\paragraph}{0pt}{*0.3}{0.5em}
\definecolor{cMHA}{HTML}{4A4A4A}
\definecolor{cGQA}{HTML}{56B4E9}
\definecolor{cMFA}{HTML}{E69F00}
\definecolor{cMLA}{HTML}{CC79A7}
\definecolor{cVHA}{HTML}{009E73}

\title{Virtual Head Attention}

\author{%
\textbf{Xibo Ding}$^{1,2}$\thanks{Equal contribution. Work done at Baidu Inc.}, ~\textbf{Guoxia Wang}$^{1}$\footnotemark[1], ~\textbf{Jinle Zeng}$^{1}$, \\
\textbf{Jiabin Yang}$^{1}$, ~\textbf{Sirui Tao}$^{2}$, 
~\textbf{Chao Yang}$^{2}$, ~\textbf{Dianhai Yu}$^{1,2}$\thanks{Corresponding author.} \\
$^1$Baidu Inc.~~$^2$Peking University \\
\texttt{\{dingxibo, wangguoxia, yudianhai\}@baidu.com}
}

\begin{document}

\maketitle

\begin{abstract}
The design of attention mechanisms in long-context large language models faces an ``impossible triangle'' among KV cache size, training and prefill FLOPs, and model quality. Grouped-Query Attention (GQA) achieves strong quality but incurs high KV cache overhead; alternative approaches such as Multi-head Latent Attention (MLA) and Multi-matrix Factorization Attention (MFA) can compress the cache, but MLA suffers from quality loss while MFA substantially increases training and prefill compute.
We propose Virtual Head Attention (VHA), which simultaneously improves all three vertices through two lightweight linear operations: Q Premix applies a near-identity transformation to queries within each KV group to recover virtual head diversity from halved physical query heads, and Linear Postmix fuses inter-head features via a low-rank $I + AB^\top$ residual structure.
Since VHA introduces no nonlinear operations, Premix can be folded into the query projection and Postmix into the output projection at inference time, reducing the model to a standard GQA-2 architecture that directly reuses all existing GQA inference optimizations. With only two KV heads, the group size in large models readily reaches 64, fully exploiting Tensor Core parallelism during decoding.
Across four scales (0.6B, 1.7B, 4B dense, and 8B-A1B MoE) with 4K-context 50B-token pretraining, VHA consistently surpasses the standard GQA baseline at every scale ($+$0.05\% to $+$0.81\%), while reducing KV cache by $4\times$ vs.\ GQA-8 at dense scales and $2\times$ vs.\ GQA-4 at the MoE scale, and lowering training and prefill FLOPs by 7.9\% (1.7B, $S{=}4096$).
On an industrial 30B-A3B MoE model trained on 500B tokens at 8K context, VHA outperforms both MLA and GQA-4 by $+$0.70\,/\,$+$2.60\,\% on a 12-benchmark base-model evaluation suite; in a separate matched 10B-A1B MoE comparison after long-context mid-training, VHA attains an average RULER score of 84.4 over 8K--128K versus 73.0 for GQA, while using two rather than four KV heads in the full-attention layers. The source code is available at \url{https://github.com/PaddlePaddle/VHA}.
\end{abstract}

\section{Introduction}

Large language models are rapidly scaling toward longer contexts, with context windows expanding from 2K tokens to 128K and beyond \citep{grattafiori2024llama, yang2024qwen2technicalreport}. As sequence lengths grow, the attention mechanism \citep{vaswani2017attention} becomes a central efficiency bottleneck in both inference and training. During inference, the key-value (KV) cache grows linearly with sequence length, consuming substantial GPU memory and constraining both serving concurrency and generation latency. During training and prefill, the FLOPs of attention computation scale quadratically with sequence length. Since the design of the attention mechanism jointly determines KV cache size, computational cost, and model expressiveness, it is a critical factor governing the overall efficiency and quality of large language models.

The design of existing attention mechanisms faces three mutually constraining objectives, forming an ``impossible triangle'': KV cache size, training and prefill FLOPs, and model quality. Consider the widely adopted Grouped-Query Attention (GQA) \citep{ainslie2023gqa}: GQA with 8 KV heads (GQA-8) achieves strong quality, but incurs a KV cache of 4\,KB per layer per token (in BF16). Directly reducing the number of KV heads to 2 (GQA-2) compresses the cache to 1\,KB, but dramatically enlarges the group size so that far more query heads must share the same key-value pairs, causing attention diversity to drop sharply and resulting in a 1.00\% quality degradation at the 1.7B scale. Multi-head Latent Attention (MLA) \citep{liu2024deepseek} applies low-rank compression to reduce KV cache to 1.12\,KB, but yields the lowest quality across multiple model scales. Multi-matrix Factorization Attention (MFA) \citep{hu2025multi} reduces the cache to 1\,KB by enlarging the head dimension, achieving quality close to GQA-8, but at the cost of a 49.8\% increase in training and prefill FLOPs (1.7B, $S{=}4096$). No existing approach achieves a satisfactory balance across all three dimensions.

We propose Virtual Head Attention (VHA), which simultaneously improves all three vertices of the impossible triangle through two lightweight linear operations. The core idea of VHA is to halve the number of physical query heads while setting the number of KV heads to only 2. Q Premix then applies a near-identity linear transformation to queries within each KV group, expanding the reduced physical query heads into virtual query heads to recover attention diversity. Linear Postmix applies a low-rank identity-residual transformation ($I + AB^\top$, $r{=}H_q'/2$, e.g., $r{=}4$ at 1.7B) on the attention output, fusing cross-group inter-head features. Since VHA introduces no nonlinear operations, Premix can be directly folded into the query projection matrix and Postmix into the output projection matrix at inference time, reducing the model to a standard GQA-2 architecture that reuses all existing GQA inference optimizations in mainstream serving frameworks such as vLLM \citep{kwon2023efficient} and SGLang \citep{zheng2024sglang}. Moreover, with only 2 KV heads, the group size in large models readily reaches 64, fully exploiting Tensor Core parallelism during the decoding phase.

Across four model scales (0.6B, 1.7B, 4B dense, and 8B-A1B MoE) with 4K-context 50B-token pretraining, VHA consistently surpasses the standard GQA baseline at every scale ($+$0.05\% to $+$0.81\%), while reducing KV cache by $4\times$ vs.\ GQA-8 at dense scales and $2\times$ vs.\ GQA-4 at the MoE scale, and lowering training and prefill FLOPs by 7.9\% (1.7B, $S{=}4096$). On an industrial 30B-A3B MoE model trained on 500B tokens at 8K context, VHA outperforms both MLA and GQA-4 by $+$0.70\,/\,$+$2.60\,\% on a 12-benchmark base-model evaluation suite, demonstrating scalability to industrial model scale. A separate matched 10B-A1B MoE comparison further extends this advantage to 128K context under more aggressive KV compression. Our main contributions are as follows:
\begin{enumerate}
    \item We propose VHA, employing Q Premix and Linear Postmix, two lightweight linear operations that simultaneously reduce KV cache and training/prefill FLOPs while improving model quality.
    \item VHA uses only linear operations and can be fully absorbed into a standard GQA-2 model at inference time, reusing existing serving frameworks such as vLLM and SGLang with zero additional overhead.
    \item Direct inspection of the trained Premix and Postmix matrices reveals two complementary structures (group-conditioned query specialization and structured low-rank cross-head communication) that jointly close the gap of the doubly-compressed configuration; component ablations confirm that neither module is sufficient on its own.
    \item Comprehensive evaluations across five scales (0.6B to 30B-A3B MoE) against MHA, GQA, MFA, and MLA establish VHA as the best attention mechanism that is Pareto-optimal on KV cache, training/prefill FLOPs, and downstream quality simultaneously; matched long-context evaluations through 128K further validate this advantage under aggressive KV compression.
\vskip -0.4cm
\end{enumerate}

\section{Related Work}

\subsection{KV Cache Efficient Attention}

Standard Multi-Head Attention (MHA) \citep{vaswani2017attention} maintains independent key and value projections per head, causing the KV cache to grow linearly with the number of heads. Multi-Query Attention (MQA) \citep{shazeer2019fast} shares a single KV head across all query heads, drastically reducing the cache at the cost of quality. Grouped-Query Attention (GQA) \citep{ainslie2023gqa} interpolates between MHA and MQA by allowing multiple query heads to share a group of KV heads, and has been widely adopted in LLaMA 2/3 \citep{touvron2023llama, grattafiori2024llama} and Qwen2 \citep{yang2024qwen2technicalreport}.

Several recent designs further compress the KV cache. Multi-head Latent Attention (MLA) \citep{liu2024deepseek} projects KV into a shared low-rank latent space and absorbs the up-projection matrices into Q and O at inference, but requires custom kernels and a non-standard inference pipeline. Multi-matrix Factorization Attention (MFA) \citep{hu2025multi} enlarges the head dimension while reducing the number of KV heads, preserving model capacity under a tight cache budget but substantially increasing training and prefill FLOPs. Tensor Product Attention (TPA) \citep{zhang2025tensor} decomposes QKV into context-dependent tensor products. Grouped-head Latent Attention (GTA) \citep{sun2025gta} extends MLA's global latent space to a group-wise design. Simulated Attention Score (SAS) \citep{zheng2025sas} projects a small number of low-dimensional KV heads into higher-dimensional spaces to simulate more attention heads, an idea related to VHA's virtual head expansion; however, SAS operates on the KV side whereas VHA virtualizes on the query side. Low-Rank Key-Value Attention (LRKV) \citep{o2026low} shares a full-rank KV basis across heads and restores head-specific flexibility through low-rank residuals, a form reminiscent of VHA's Postmix ($I+AB^\top$); yet LRKV acts in KV space and its additional structure remains at inference time. Cross-Layer Attention (CLA) \citep{brandon2024reducing} and YOCO \citep{sun2024you} reduce memory by sharing or reusing KV caches across layers.

Unlike all of the above, VHA employs a richer training-time parameterization that merges exactly back into standard GQA at deployment, requiring no custom kernels or specialized inference procedures.

\vskip -0.4cm
\subsection{Inter-Head Communication and Virtual Heads}

Talking-Heads Attention \citep{shazeer2020talking} applies learned linear projections across heads before and after softmax, pioneering inter-head communication. \citet{cordonnier2020multi} analyze multi-head attention from a collaborative perspective, revealing substantial redundancy among heads. More recently, Explicit Multi-head Attention (MEA) \citep{peng2026explicit} introduces a head-level linear composition module for explicit cross-head interaction, and Interleaved Head Attention \citep{duvvuri2026interleaved} interleaves head representations to enable information exchange. In contrast to these methods, VHA does not mix across heads in the attention weight space; instead, it virtualizes in the query space via Premix and fuses inter-head features in the output space via a low-rank Postmix, with all operations being strictly linear and fully absorbable at inference.

\vskip -0.4cm
\subsection{Reparameterization}

RepVGG \citep{ding2021repvgg} uses a multi-branch training architecture ($3{\times}3 + 1{\times}1 + \text{identity}$) that merges into a single convolution at inference, demonstrating that decoupled training-time parameterization provides implicit regularization at zero inference cost. LoRA \citep{hu2022lora} performs parameter-efficient fine-tuning via low-rank decomposition ($W + BA$). VHA extends this reparameterization paradigm from a single weight matrix to the head dimension of attention: Premix structurally factorizes the per-virtual-head query projections ($\widetilde{W}_Q^{(i,g)} = W_Q^{(i)} W_g^{\mathrm{premix}}$), while Postmix introduces a low-rank identity-residual mixing on the head axis ($I + AB^\top$). Both modules are mathematically equivalent in the forward pass to a standard GQA configuration, yet the decoupled parameterization yields distinct optimization dynamics during training and merges back into a standard GQA-2 architecture at inference with zero overhead.

\vskip -0.4cm
\section{Method: Virtual Head Attention}

\begin{figure*}[t]
\centering
\includegraphics[width=\textwidth]{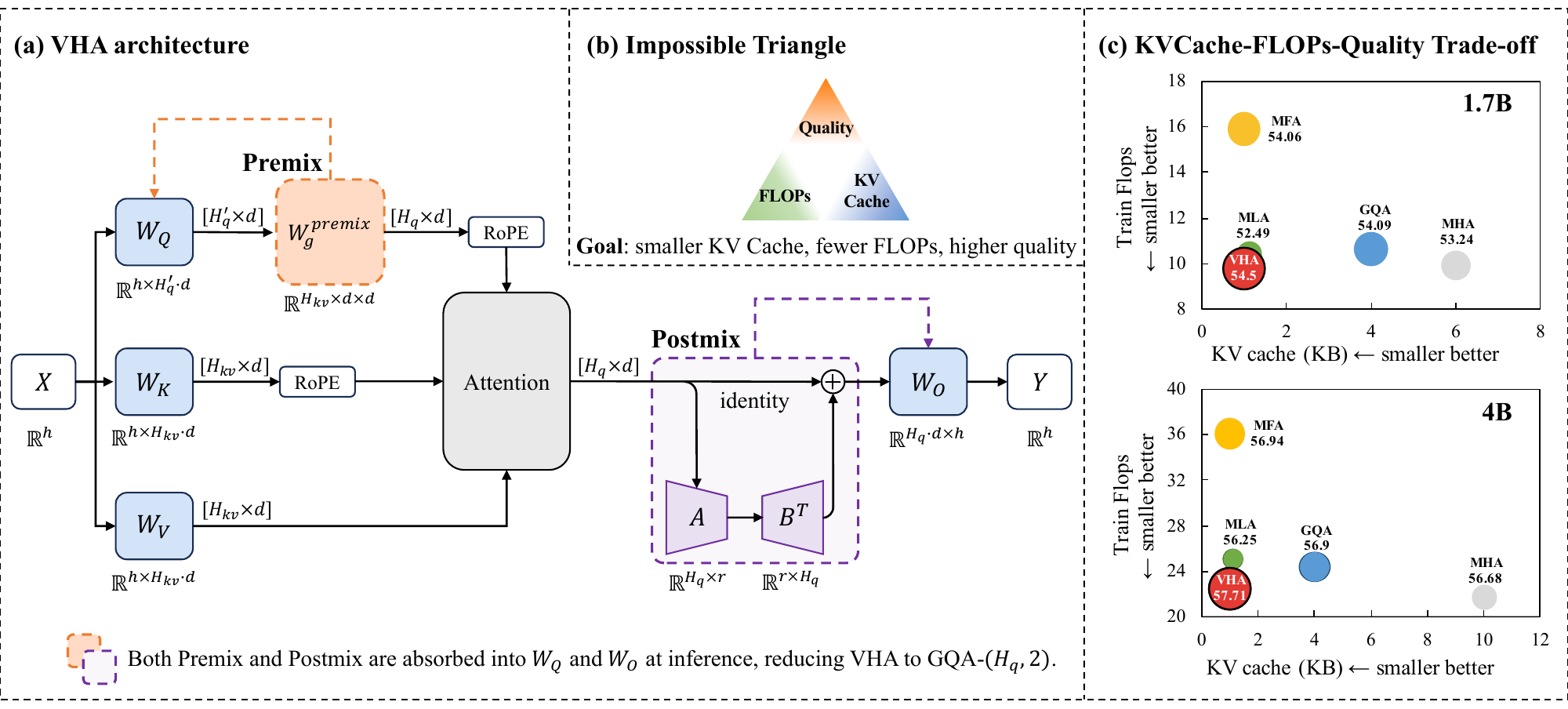}
\vskip -0.1cm
\caption{\textbf{Overview of Virtual Head Attention.} (a)~VHA improves the trade-off among KV cache size, training/prefill FLOPs, and model quality by combining Q Premix and Linear Postmix during training, while absorbing both linear modules into standard GQA projections at inference. (b)~Schematic of the KV cache--FLOPs--quality impossible triangle faced by existing attention designs. (c)~Method comparison at 1.7B and 4B scales, plotting KV cache size against training FLOPs, with bubble area proportional to 8-task average accuracy.}
\label{fig:overview}
\vskip -0.4cm
\end{figure*}

\vskip -0.1cm
\subsection{Preliminaries}

Grouped-Query Attention (GQA) \citep{ainslie2023gqa} partitions $H_q$ query heads into $H_{kv}$ groups, where heads within each group share a common key-value pair. Given input $X \in \mathbb{R}^{S \times h}$, GQA computes $Q = XW_Q$, $K = XW_K$, $V = XW_V$, where $W_Q \in \mathbb{R}^{h \times H_q d}$, $W_K \in \mathbb{R}^{h \times H_{kv} d}$, $W_V \in \mathbb{R}^{h \times H_{kv} d}$. Each query head attends to its group's key-value pair via scaled dot-product attention \citep{vaswani2017attention}, and all head outputs are concatenated and projected by $W_O \in \mathbb{R}^{H_q d \times h}$. During inference, K and V are cached, consuming $4H_{kv}d$ bytes per layer per token in BF16. Reducing $H_{kv}$ compresses the cache but enlarges the group size $G = H_q / H_{kv}$, forcing more query heads to share the same key-value pair and degrading attention diversity.

\subsection{Overall Design}

VHA halves the number of physical query heads while setting $H_{kv} = 2$, then recovers model expressiveness through two lightweight linear operations. Specifically, $H_q' = H_q / 2$ physical query heads and $H_{kv} = 2$ KV heads yield $H_v = H_{kv} \times H_q' = H_q$ virtual heads (matching the original GQA query head count), with $W_O \in \mathbb{R}^{H_v d \times h}$ identical in size to the original GQA output projection. The resulting KV cache is $4 \times 2 \times d$ bytes per layer per token, $4\times$ smaller than GQA-8 at dense scales and $2\times$ smaller than GQA-4 at MoE scales.

On top of this configuration, VHA introduces \textbf{Q~Premix}, which expands $H_q'$ physical query heads into $H_v$ virtual query heads to restore attention diversity, and \textbf{Linear Postmix}, which applies a low-rank identity-residual transformation on the attention output to fuse inter-head features.

\vskip -0.1cm
\paragraph{Inference absorption principle.} VHA uses only linear operations (no nonlinear activations), so that at inference time Premix can be folded into $W_Q$ and Postmix into $W_O$, reducing the entire model to a standard GQA-2 architecture that directly reuses all existing GQA optimizations in frameworks such as vLLM \citep{kwon2023efficient} and SGLang \citep{zheng2024sglang}.

\subsection{Q Premix: Virtual Query Head Expansion}

Q Premix applies a near-identity linear transformation to physical query heads within each KV group, expanding them into virtual query heads. For KV group $g \in \{1, \dots, H_{kv}\}$, each physical query head $Q_i \in \mathbb{R}^{S \times d}$, $i \in \{1, \dots, H_q'\}$, is transformed as:
\begin{equation}
    \tilde{Q}_{i,g} = Q_i \, W_g^{\mathrm{premix}}, \quad W_g^{\mathrm{premix}} \in \mathbb{R}^{d \times d}
\end{equation}
where $W_g^{\mathrm{premix}}$ is a per-group transformation matrix. We initialize $W_g^{\mathrm{premix}} = I + \alpha \cdot \epsilon$, $\epsilon \sim \mathcal{N}(0, \sigma^2)$, with a small scaling factor $\alpha$. This near-identity initialization ensures that virtual query heads start as slight perturbations of the physical query heads, preserving training stability at the early stage. As training progresses, each $W_g^{\mathrm{premix}}$ gradually deviates from the identity to produce increasingly diverse attention patterns across virtual heads. Through this operation, $H_q'$ physical query heads are expanded into $H_v = H_{kv} \times H_q'$ virtual query heads.

\vskip -0.1cm
\paragraph{Structured factorization view.} Premix is mathematically equivalent in the forward pass to a standard $H_v$-head GQA whose query projections are structurally factorized. Specifically, defining
\begin{equation}
    \widetilde{W}_Q^{(i,g)} \;=\; W_Q^{(i)} \, W_g^{\mathrm{premix}} \;\in\; \mathbb{R}^{h \times d},
\end{equation}
the entire Premix-then-attend computation reduces, for any input $X$ and any parameter values, to
\begin{equation}
    \mathrm{softmax}\!\left(\tilde{Q}_{i,g} K_g^\top / \sqrt{d}\right) V_g \;=\; \mathrm{softmax}\!\left(X \widetilde{W}_Q^{(i,g)} K_g^\top / \sqrt{d}\right) V_g.
\end{equation}
Thus VHA is a standard GQA with $H_v = H_{kv} H_q'$ virtual query heads and $H_{kv}$ KV heads, where the per-virtual-head query projections are constrained to lie on the structured low-dimensional manifold $\{ W_Q^{(i)} W_g^{\mathrm{premix}} \}$.

\vskip -0.1cm
\paragraph{Why this factorization is useful.} The structured parameterization brings two benefits over an unconstrained $H_v$-head GQA. First, it is parameter- and compute-efficient: the Q path uses $H_q' h d + H_{kv} d^2$ parameters and a comparable number of FLOPs, versus $H_v h d$ for a fully decoupled $H_v$-head Q projection. At the 1.7B scale, this yields a roughly 50\% reduction in Q-projection parameters and FLOPs (the primary source of VHA's training-FLOPs reduction). Second, although the forward pass is algebraically equivalent to training a single $\widetilde{W}_Q^{(i,g)}$, the gradients of $W_Q^{(i)}$ and $W_g^{\mathrm{premix}}$ are coupled across all groups and physical heads, yielding a different optimization landscape and providing implicit regularization analogous to LoRA \citep{hu2022lora} and RepVGG \citep{ding2021repvgg} (see Appendix~\ref{app:premix_proof} for a formal analysis).

\vskip -0.1cm
\paragraph{Inference-time absorption.} Once training is complete and parameters are frozen, for each virtual head $(i, g)$ we precompute $\widetilde{W}_Q^{(i,g)} = W_Q^{(i)} \cdot W_g^{\mathrm{premix}}$, after which VHA reduces to a standard GQA-2 with $H_v$ independent query projections and $H_{kv}=2$ KV heads.

\subsection{Linear Postmix: Low-Rank Identity-Residual Mixing}
\label{sec:postmix}

While Premix restores virtual head diversity within each KV group, the outputs across different groups still lack cross-group information exchange. Linear Postmix addresses this through a low-rank identity-residual structure.

Let $Y \in \mathbb{R}^{S \times H_v \times d}$ denote the attention output. Postmix computes:
\begin{equation}
    Y' = Y + B A^\top Y = \left(I + BA^\top\right)Y
\end{equation}
where $A, B \in \mathbb{R}^{H_v \times r}$ and $r$ is the bottleneck rank (default $r = H_q'/2$, e.g., $r{=}4$ at 1.7B). The operation acts along the head dimension independently for each token and head-dimension channel, applying the transformation $I + BA^\top \in \mathbb{R}^{H_v \times H_v}$ to the head axis.

\vskip -0.1cm
\paragraph{Initialization.} $A$ is initialized with small random values and $B$ with zeros, ensuring $BA^\top = 0$ at the start of training so that Postmix acts as the identity and does not disturb the initial training dynamics. As training progresses, $B$ gradually learns effective cross-group mixing patterns.

\vspace{-0.1cm}
\paragraph{Linear redundancy and reparameterization benefit.} Mathematically, Postmix can be absorbed into $W_O$. Using the row-major Kronecker identity $\mathrm{vec}_\text{row}(MY)^\top = \mathrm{vec}_\text{row}(Y)^\top (M^\top \otimes I_d)$ \citep{roger1994topics}:
\begin{equation}
    \mathrm{vec}_\text{row}(Y)^\top \!\left((I_{H_v} + AB^\top) \otimes I_d\right) W_O = \mathrm{vec}_\text{row}(Y)^\top W_O'
\end{equation}
where $W_O' = \bigl((I_{H_v} + AB^\top) \otimes I_d\bigr)\, W_O \in \mathbb{R}^{H_v d \times h}$. Thus, linear Postmix is theoretically redundant (see Appendix~\ref{app:postmix_proof} for the complete derivation). However, ablation experiments (Section~\ref{sec:ablations}) show that adding Postmix on top of Premix yields a $+$1.19\,\% improvement at 1.7B, analogous to the implicit regularization observed in RepVGG \citep{ding2021repvgg} and the low-rank decomposition in LoRA \citep{hu2022lora}: the separated gradient paths of $A$ and $B$ provide a different optimization landscape, while the low-rank constraint ($r \ll H_v$) acts as an additional regularizer preventing cross-head mixing from overfitting.

\vspace{-0.1cm}
\paragraph{Inference-time absorption.} At inference, we precompute $W_O' = \bigl((I_{H_v} + AB^\top) \otimes I_d\bigr) W_O$, after which VHA reduces to standard GQA-2. Algorithm~\ref{alg:vha} (Appendix~\ref{app:algorithm}) summarizes the complete training forward pass and the one-time inference-absorption procedure.

\subsection{Efficiency Analysis}

\paragraph{Training and prefill FLOPs.} By halving the physical $H_q$, VHA directly reduces Q projection FLOPs. The additional cost of Premix ($2 H_{kv} H_q' d^2$) and Postmix ($4 H_v r d$) is negligible compared to the FLOPs saved by halving the Q projection. Table~\ref{tab:efficiency} quantifies this trade-off in detail across all four model scales ($S{=}4096$); VHA emerges as the best Pareto-optimal point on KV cache and FLOPs at every scale (Section~\ref{sec:efficiency}).

\paragraph{Decoding inference.} Autoregressive decoding is memory-bound \citep{pope2023efficiently}, so KV cache size directly governs latency and serving concurrency. With KV cache reduced by $4\times$ vs.\ GQA-8 (dense) or $2\times$ vs.\ GQA-4 (MoE), VHA achieves a corresponding 2--4$\times$ theoretical speedup in memory-bound decoding. Since Premix and Postmix are fully absorbed at inference, VHA's decoding cost is identical to standard GQA-2.
\vskip -0.4cm

\begin{table*}[t]
\centering
\footnotesize
\caption{\textbf{Pretraining accuracy across four model scales.}
We report zero-shot accuracy (\%) on eight standard multiple-choice benchmarks.
\textsc{Cache} is KV cache per layer per token in BF16 (KB).
\textbf{Bold} marks the best per scale per column; ties are bolded jointly.
VHA achieves the highest average accuracy at every scale while using the
smallest KV cache. Per-method full configurations and benchmark citations
are in Appendix~\ref{app:setup}; per-method training-FLOPs comparison is in Table~\ref{tab:efficiency}.}
\label{tab:main-compare}
\setlength{\tabcolsep}{2.8pt}
\renewcommand{\arraystretch}{1.05}
\scriptsize
\centering
\resizebox{\textwidth}{!}{%
\begin{tabular}{llcccccccccccc}
\toprule
Scale & Attn & Q-K-V & Cache & ARC-C & BoolQ & HSwag & OBQA & PIQA & Wino & SciQ & SIQA & \textbf{Avg} \\
\midrule
\multirow{5}{*}{0.6B}
 & MHA & 12-12-12 & 6.00          & 29.10 & 60.40 & 44.88 & 33.20 & 68.66 & 51.85 & 74.50 & 39.15 & 50.22 \\
 & GQA & 16-8-8   & 4.00          & 29.86 & 58.01 & 44.77 & \textbf{35.80} & 68.17 & 52.33 & 75.70 & 38.69 & 50.42 \\
 & MLA & 14-14-14 & 1.12          & 29.27 & 55.96 & 43.20 & 33.80 & 68.23 & 50.67 & \textbf{76.40} & 38.59 & 49.52 \\
 & MFA & 17-1-1   & \textbf{1.00} & 30.80 & 60.18 & \textbf{45.47} & 32.80 & \textbf{69.04} & \textbf{54.70} & 73.20 & 39.05 & 50.66 \\
 & \textbf{VHA} & 8-2-2 & \textbf{1.00} & \textbf{31.91} & \textbf{61.19} & 45.26 & 35.60 & 68.44 & 50.99 & 75.60 & \textbf{39.36} & \textbf{51.04} \\
\midrule
\multirow{5}{*}{1.7B}
 & MHA & 12-12-12 & 6.00          & 35.24 & 58.53 & 51.86 & 35.20 & 71.33 & 54.06 & 80.30 & 39.36 & 53.24 \\
 & GQA & 16-8-8   & 4.00          & 35.15 & \textbf{62.11} & 51.98 & 35.60 & 71.71 & \textbf{55.25} & 81.10 & 39.82 & 54.09 \\
 & MLA & 12-12-12 & 1.12          & 31.57 & 61.31 & 49.22 & 36.20 & 70.51 & 54.22 & 77.60 & 39.30 & 52.49 \\
 & MFA & 19-1-1   & \textbf{1.00} & 34.47 & 59.45 & \textbf{52.16} & 37.80 & \textbf{71.82} & 54.85 & 81.50 & 40.43 & 54.06 \\
 & \textbf{VHA} & 8-2-2 & \textbf{1.00} & \textbf{35.92} & 60.64 & 51.71 & \textbf{39.00} & 71.38 & 54.70 & \textbf{81.70} & \textbf{40.94} & \textbf{54.50} \\
\midrule
\multirow{5}{*}{4B}
 & MHA & 20-20-20 & 10.00         & 38.31 & 60.21 & 57.52 & 39.60 & 73.88 & 56.99 & \textbf{85.40} & 41.56 & 56.68 \\
 & GQA & 32-8-8   & 4.00          & \textbf{39.16} & 62.32 & 58.39 & 39.40 & 73.83 & 57.06 & 83.80 & 41.20 & 56.90 \\
 & MLA & 28-28-28 & 1.12          & 37.20 & 61.65 & 56.55 & 38.80 & 73.29 & 56.75 & 84.90 & 41.86 & 56.38 \\
 & MFA & 33-1-1   & \textbf{1.00} & 38.40 & 62.20 & 57.67 & \textbf{39.80} & 73.88 & 57.70 & 83.60 & 42.27 & 56.94 \\
 & \textbf{VHA} & 16-2-2 & \textbf{1.00} & 38.91 & \textbf{64.74} & \textbf{58.67} & 39.20 & \textbf{74.43} & \textbf{58.80} & 84.60 & \textbf{42.34} & \textbf{57.71} \\
\midrule
\multirow{5}{*}{8B-A1B}
 & MHA & 20-20-20 & 10.00         & 35.67 & 59.63 & 53.98 & 37.40 & 72.25 & 53.04 & \textbf{82.70} & 42.12 & 54.60 \\
 & GQA & 32-4-4   & 2.00          & 34.73 & 61.53 & 54.03 & \textbf{38.60} & 71.55 & 55.09 & 82.60 & 40.74 & 54.86 \\
 & MLA & 28-28-28 & 1.12          & 33.11 & 60.86 & 52.29 & 36.60 & 72.63 & 53.28 & 82.00 & 39.56 & 53.79 \\
 & MFA & 33-1-1   & \textbf{1.00} & 35.07 & \textbf{62.48} & \textbf{54.08} & 36.60 & \textbf{72.80} & \textbf{55.17} & 81.80 & 40.48 & 54.81 \\
 & \textbf{VHA} & 16-2-2 & \textbf{1.00} & \textbf{36.69} & 62.17 & 53.78 & 37.00 & 71.71 & 54.06 & 82.40 & \textbf{41.45} & \textbf{54.91} \\
\bottomrule
\end{tabular}%
}
\vskip -0.5cm
\end{table*}

\section{Experiments}
\label{sec:experiments}

\vskip -0.2cm
\subsection{Experimental Setup}
\label{sec:setup}

\paragraph{Baselines.} We compare VHA with MHA, GQA, MFA, and MLA across all scales. Following common practice, GQA uses 8 KV heads at dense scales (0.6B/1.7B/4B) and 4 KV heads at MoE scales (8B-A1B and 30B-A3B). VHA halves the physical query heads relative to GQA and uses $H_{kv}{=}2$, yielding $4\times$ KV cache compression at dense scales and $2\times$ at MoE scales.
\vskip -0.4cm

\paragraph{Small-scale pretraining.} We pretrain Qwen3-architecture models at four scales (dense at 0.6B, 1.7B, and 4B parameters, plus an 8B-A1B MoE) on 50B tokens of FineWeb-Edu with BF16 mixed precision and a 4K context length. We report zero-shot accuracy on eight standard benchmarks.
\vskip -0.4cm

\paragraph{Industrial-scale pretraining.} We pretrain an ERNIE-style 30B-A3B MoE on an internal multi-domain pretraining corpus (web text, books, code, and multilingual data) with FP8 mixed precision on 500B tokens at 8K context. We evaluate on 12 standard base-model benchmarks spanning general knowledge, math \& STEM, coding, and multilingual capabilities. Full model configurations, benchmark details, and training hyperparameters are in Appendix~\ref{app:setup}.

\paragraph{Long-context mid-training.} We additionally compare VHA and GQA on a matched MiMo-2.5-Flash-style 10B-A1B MoE backbone that interleaves full and sliding-window attention at a 1:3 ratio. The two models share the backbone, data, optimization, and sliding-window layers; only the full-attention layers differ, using two KV heads for VHA and four for GQA. We evaluate through 128K, with the stage-wise recipe and benchmark definitions reported in Appendix~\ref{app:long-context}.

\subsection{Main Results}
\vskip -0.2cm
\label{sec:main-results}

\paragraph{Cross-scale consistency.} VHA achieves the highest average accuracy at every scale, surpassing the GQA baseline by $+$0.05\% to $+$0.81\% while using the smallest KV cache: 1.00\,KB per layer per token, tied with MFA, $4\times$ smaller than GQA-8 at dense scales and $2\times$ smaller than GQA-4 at the MoE scale. The advantage holds across both dense and MoE architectures. A separate matched-harness study at 0.6B confirms the same ordering across repeated runs, with VHA reaching $50.40{\pm}0.07$ ($n{=}4$) versus GQA's $50.00{\pm}0.34$ ($n{=}3$); full results are in Appendix~\ref{app:robustness}.

\paragraph{Trade-offs against each baseline.} At equal KV cache (1.00\,KB), VHA dominates MFA at every scale ($+$0.10\% to $+$0.77\%) while reducing training and prefill FLOPs by 7.9\%, compared with MFA's $+$49.8\% increase at the 1.7B scale ($S{=}4096$; Table~\ref{tab:efficiency}). Full-cache MHA falls below GQA-8 at every dense scale despite using a $\geq 1.5\times$ larger cache, confirming that grouping the queries is essentially free. MLA underperforms at every scale, suggesting that latent KV compression is less effective than head-level virtualization at the 50B-token training budget used here. Per-method training loss curves at all four scales are reported in Appendix~\ref{app:loss-curves}.

\begin{table*}[t]
\centering
\footnotesize
\caption{\textbf{Recovering from a doubly-compressed shape (1.7B).}
GQA 16-8-8 is the standard GQA-8 baseline; GQA 16-2-2 reduces only $H_{kv}$
to obtain a 1\,KB KV cache; GQA 8-2-2 additionally halves the physical query
heads to match VHA's shape but lacks Premix and Postmix. $\Delta$ is the
gap to the GQA 16-8-8 baseline.}
\label{tab:doubly-compressed}
\setlength{\tabcolsep}{3pt}
\renewcommand{\arraystretch}{1.05}
\scriptsize
\resizebox{\textwidth}{!}{%
\begin{tabular}{lcccccccccc r}
\toprule
Method & Cache & ARC-C & BoolQ & HSwag & OBQA & PIQA & Wino & SciQ & SIQA & \textbf{Avg} & $\Delta$ \\
\midrule
GQA 16-8-8         & 4.00          & 35.15 & \textbf{62.11} & \textbf{51.98} & 35.60 & \textbf{71.71} & \textbf{55.25} & 81.10 & 39.82 & 54.09 & --- \\
GQA 16-2-2         & \textbf{1.00} & 33.19 & 58.96 & 50.92 & 36.00 & 71.27 & 54.22 & 79.60 & 40.53 & 53.09 & $-$1.00 \\
GQA 8-2-2          & \textbf{1.00} & 33.02 & 59.11 & 49.98 & 34.00 & 71.44 & 53.28 & 77.80 & 40.43 & 52.38 & $-$1.71 \\
\textbf{VHA 8-2-2} & \textbf{1.00} & \textbf{35.92} & 60.64 & 51.71 & \textbf{39.00} & 71.38 & 54.70 & \textbf{81.70} & \textbf{40.94} & \textbf{54.50} & $+$0.41 \\
\bottomrule
\end{tabular}%
}
\vskip -0.5cm
\end{table*}

\paragraph{Recovering from a doubly-compressed shape.}
\label{sec:doubly-compressed}
Reducing GQA's KV head count from 8 to 2 (GQA 16-2-2) costs $-$1.00\% in average accuracy at the 1.7B scale; further halving the physical query heads to match VHA's shape (GQA 8-2-2) costs another $-$0.71\%, totaling $-$1.71\% vs.\ the GQA-8 baseline. With the same (8, 2, 2) shape, VHA recovers all of this loss and exceeds the original baseline by $+$0.41\%, despite using only 25\% of GQA-8's KV cache and 50\% of its physical query heads. The doubly-compressed (8, 2, 2) shape that on its own forfeits $-$1.71\% thus becomes a winning configuration once Premix and Postmix are introduced.

\vskip -0.4cm
\subsection{Large-scale and long-context validation}
\label{sec:large-scale}

\begin{table*}[t]
\centering
\footnotesize
\caption{\textbf{Pretraining accuracy at the 30B-A3B MoE scale.}
We report accuracy (\%) averaged within four task categories
(General, Math \& STEM, Coding, Multilingual) over 12 base-model
benchmarks; \textsc{Avg} is the macro-average over the four task categories.
\textsc{Cache} is KV cache per layer per token in BF16 (KB).
\textbf{Bold} marks the best per column.
Per-benchmark scores and benchmark citations are in Appendix~\ref{app:per-benchmark}.}
\label{tab:large-scale}
\setlength{\tabcolsep}{4pt}
\renewcommand{\arraystretch}{1.05}
\footnotesize
\centering
\begin{tabular}{llcccccc}
\toprule
Method & Q-K-V & Cache & General & Math \& STEM & Coding & Multilingual & \textbf{Avg} \\
\midrule
GQA              & 48-4-4   & 2.00          & 52.86 & 43.09 & 51.85 & 43.26 & 47.76 \\
MLA              & 28-28-28 & 1.12          & 55.57 & 43.49 & \textbf{54.37} & 45.20 & 49.66 \\
\textbf{VHA}     & 24-2-2   & \textbf{1.00} & \textbf{56.25} & \textbf{46.40} & 53.23 & \textbf{45.56} & \textbf{50.36} \\
\bottomrule
\end{tabular}
\vskip -0.5cm
\end{table*}

At the 30B-A3B MoE scale (500B tokens; FP8 mixed precision; full setup in Appendix~\ref{app:setup}), Table~\ref{tab:large-scale} shows that VHA 24-2-2 attains the highest 12-benchmark average while using only 50\% of GQA-4's KV cache and 89\% of MLA's, confirming that the trends observed at 0.6B$\sim$8B-A1B carry over to the production-scale MoE regime. The category-level breakdown shows that gains are broadly distributed: General $+$3.39\,/\,$+$0.68\%, Math\,\&\,STEM $+$3.31\,/\,$+$2.91\%, Coding $+$1.38\,/\,$-$1.14\%, Multilingual $+$2.30\,/\,$+$0.36\% (VHA vs.\ GQA\,/\,MLA), indicating head-level virtualization scales gracefully across heterogeneous downstream tasks. Per-benchmark scores for all 12 evaluations are in Appendix~\ref{app:per-benchmark}.

\paragraph{Long-context validation.} At the final 128K checkpoint, VHA outperforms the matched GQA baseline on RULER at every evaluated length from 8K to 128K, with a length-average of 84.4 versus 73.0 ($+$11.4 points). The advantage extends to complementary evaluations at 128K: VHA outperforms GQA on LongBench-v2 (24.2 vs.\ 18.9) and BABILong (38.2 vs.\ 31.1), while remaining on par on InfiniteBench (31.5 vs.\ 31.8); averaged across lengths, VHA wins, ties, and loses 3/1/1 HELMET categories. Together, these results demonstrate that VHA preserves its quality advantage through 128K while halving the KV heads and KV cache of the full-attention layers relative to GQA-4; complete results are in Appendix~\ref{app:long-context}.

\subsection{Ablations}
\label{sec:ablations}

We ablate VHA's design choices at the 1.7B scale (additionally at 4B for the component contribution), isolating (i) the contribution of Premix and Postmix, (ii) the Postmix rank, and (iii) whether non-linearity helps either module.

Reading the $\Delta$ column of Table~\ref{tab:ablation} (left and middle), the two modules address complementary deficits of the doubly-compressed shape: only their joint VHA configuration consistently achieves the largest gains at both scales at $25\%$ of GQA-8's KV cache; we provide a mechanistic interpretation of this complementarity in Section~\ref{sec:analysis}.

\begin{table*}[t]
\centering
\footnotesize
\caption{\textbf{VHA ablations.} \emph{Left, middle:} component
contribution at 1.7B and 4B (each row uses VHA's shape, 8-2-2 or
16-2-2, with KV cache 1.00\,KB / layer / token vs 4.00\,KB for the
GQA-8 baseline; Premix and Postmix are toggled individually).
\emph{Right:} activation ablation at 1.7B (replacing the linear
formulation with SiLU on either or both modules). Avg (\%) is over
the eight benchmarks of Table~\ref{tab:main-compare}; $\Delta$ is the
average delta vs the GQA-8 baseline (component) or the linear default
(activation); the linear VHA default is \textbf{bold}. Each module on
its own only partially recovers the doubly-compressed loss from
Table~\ref{tab:doubly-compressed}; the full VHA configuration exceeds
the GQA-8 baseline at both scales, and SiLU consistently underperforms
the linear default.}
\label{tab:ablation}
\setlength{\tabcolsep}{4pt}
\renewcommand{\arraystretch}{1.05}
\begin{minipage}[t]{0.31\textwidth}
\centering
\textbf{Component @ 1.7B}\\[0.3em]
\begin{tabular}{lcc}
\toprule
Variant       & Avg            & $\Delta$ \\
\midrule
GQA           & 54.09          & ---             \\
Premix only   & 53.31          & $-$0.78         \\
Postmix only  & 53.95          & $-$0.14         \\
\textbf{VHA}  & \textbf{54.50} & \textbf{$+$0.41}\\
\bottomrule
\end{tabular}
\end{minipage}
\hfill
\begin{minipage}[t]{0.31\textwidth}
\centering
\textbf{Component @ 4B}\\[0.3em]
\begin{tabular}{lcc}
\toprule
Variant       & Avg            & $\Delta$ \\
\midrule
GQA           & 56.90          & ---             \\
Premix only   & 56.62          & $-$0.28         \\
Postmix only  & 57.22          & $+$0.32         \\
\textbf{VHA}  & \textbf{57.71} & \textbf{$+$0.81}\\
\bottomrule
\end{tabular}
\end{minipage}
\hfill
\begin{minipage}[t]{0.31\textwidth}
\centering
\textbf{Activation @ 1.7B}\\[0.3em]
\begin{tabular}{lcc}
\toprule
Variant                   & Avg            & $\Delta$ \\
\midrule
+ SiLU on Premix          & 53.83          & $-$0.67  \\
+ SiLU on Postmix         & 53.79          & $-$0.71  \\
+ SiLU on both            & 53.67          & $-$0.83  \\
\textbf{Linear (default)} & \textbf{54.50} & ---      \\
\bottomrule
\end{tabular}
\end{minipage}
\vskip -0.5cm
\end{table*}

\paragraph{Postmix rank.} With Premix fixed at the 1.7B scale, sweeping the Postmix rank gives Avg $54.50 / 54.24 / 54.15$ for $r \in \{4, 32, 128\}$: larger $r$ monotonically lowers training loss but degrades downstream evaluation, indicating that the low-rank constraint provides useful implicit regularization. We therefore fix $r = H_q'/2$ across all scales (Section~\ref{sec:postmix} / Appendix~\ref{app:setup}).

\paragraph{Activation.} Replacing the linear formulation with SiLU on either or both modules consistently degrades quality at 1.7B (Table~\ref{tab:ablation}, right), regardless of where the non-linearity is inserted. Beyond the empirical loss, every non-linear variant forfeits inference-time absorption: with both modules linear the entire VHA computation collapses to standard GQA-2 at serving time, whereas any non-linearity must remain in the runtime kernel. We therefore keep both modules strictly linear, simultaneously achieving the highest empirical quality and lossless absorption into standard GQA-2 at inference.

\subsection{Computational Efficiency}
\label{sec:efficiency}

\begin{table*}[t]
\centering
\caption{\textbf{Computational efficiency across four model scales} ($S{=}4096$).
For each scale we list per-method query/key/value head counts, KV cache
per layer per token (BF16, KB), training FLOPs per token aggregated over
all transformer layers (G), and the relative change vs.\ the per-scale
GQA baseline (GQA-8 dense, GQA-4 MoE). \textbf{Bold} highlights VHA, the method that achieves the best joint KV cache$+$FLOPs Pareto optimum at every
scale. MHA's lower FLOPs at three scales reflect its smaller query head
count under our parameter-matched configurations ($H_q{=}12$ at 0.6B/1.7B,
$H_q{=}20$ at 4B/8B-A1B versus GQA's $H_q{=}16$ or $H_q{=}32$), at the
cost of a $6$--$10\times$ larger KV cache.}
\label{tab:efficiency}
\setlength{\tabcolsep}{2pt}
\renewcommand{\arraystretch}{1.05}
\scriptsize
\resizebox{\textwidth}{!}{%
\begin{tabular}{l cccc cccc cccc cccc}
\toprule
& \multicolumn{4}{c}{\textbf{0.6B}}
& \multicolumn{4}{c}{\textbf{1.7B}}
& \multicolumn{4}{c}{\textbf{4B}}
& \multicolumn{4}{c}{\textbf{8B-A1B}} \\
\cmidrule(lr){2-5} \cmidrule(lr){6-9} \cmidrule(lr){10-13} \cmidrule(lr){14-17}
Method
& Q-K-V & Cache & FLOPs & $\Delta$\,\%
& Q-K-V & Cache & FLOPs & $\Delta$\,\%
& Q-K-V & Cache & FLOPs & $\Delta$\,\%
& Q-K-V & Cache & FLOPs & $\Delta$\,\% \\
\midrule
MHA          & 12-12-12       & 6.00          & 6.87          & $-$9.3          & 12-12-12       & 6.00          & 9.91          & $-$6.6          & 20-20-20         & 10.00         & 21.66          & $-$11.1         & 20-20-20         & 10.00         & 8.76           & $-$14.7         \\
GQA          & 16-8-8         & 4.00          & 7.57          & 0               & 16-8-8         & 4.00          & 10.61         & 0               & 32-8-8           & 4.00          & 24.38          & 0               & 32-4-4           & 2.00          & 10.27          & 0               \\
MLA          & 14-14-14       & 1.12          & 8.25          & $+$8.9          & 12-12-12       & 1.12          & 10.42         & $-$1.8          & 28-28-28         & 1.12          & 25.09          & $+$2.9          & 28-28-28         & 1.12          & 11.39          & $+$10.9         \\
MFA          & 17-1-1         & 1.00          & 11.54         & $+$52.3         & 19-1-1         & 1.00          & 15.90         & $+$49.8         & 33-1-1           & 1.00          & 36.04          & $+$47.9         & 33-1-1           & 1.00          & 17.82          & $+$73.5         \\
\textbf{VHA} & \textbf{8-2-2} & \textbf{1.00} & \textbf{6.96} & \textbf{$-$8.1} & \textbf{8-2-2} & \textbf{1.00} & \textbf{9.78} & \textbf{$-$7.9} & \textbf{16-2-2}  & \textbf{1.00} & \textbf{22.52} & \textbf{$-$7.6} & \textbf{16-2-2}  & \textbf{1.00} & \textbf{9.60}  & \textbf{$-$6.5} \\
\bottomrule
\end{tabular}%
}
\vskip -0.5cm
\end{table*}

Table~\ref{tab:efficiency} quantifies the joint KV cache--FLOPs trade-off across the four model scales of Table~\ref{tab:main-compare}. VHA is the only method that simultaneously matches the smallest KV cache (1.00\,KB) and reduces training/prefill FLOPs below the per-scale GQA baseline ($-$8.1\,/\,$-$7.9\,/\,$-$7.6\,/\,$-$6.5\,\% across scales). MFA matches the cache but pays a $+$48--74\,\% FLOPs penalty; MLA's latent compression yields a slightly larger cache (1.12\,KB) and raises FLOPs at three of four scales (up to $+$10.9\,\% at 8B-A1B). MHA appears competitive on FLOPs only because its smaller query head count under our parameter-matched configurations comes with a $6$--$10\times$ larger KV cache. Combined with VHA's complete inference absorption to standard GQA-2 (Section~\ref{sec:postmix}), this makes VHA the best Pareto-optimal point on the cache--FLOPs--quality triangle in our experiments.

\section{Analysis: How VHA Reparameterization Works}
\vskip -0.1cm
\label{sec:analysis}

Beyond task accuracy, VHA should exploit the limited information available under aggressive KV sharing. We therefore inspect the Premix and Postmix matrices of trained VHA models (Premix on the 1.7B, Postmix on the 4B model), asking whether (i)~Premix learns a group-conditioned query geometry, (ii)~Postmix learns structured low-rank cross-head communication, and (iii)~the two jointly close the doubly-compressed gap of Section~\ref{sec:ablations}.

\begin{figure}[t]
\centering
\includegraphics[width=\linewidth]{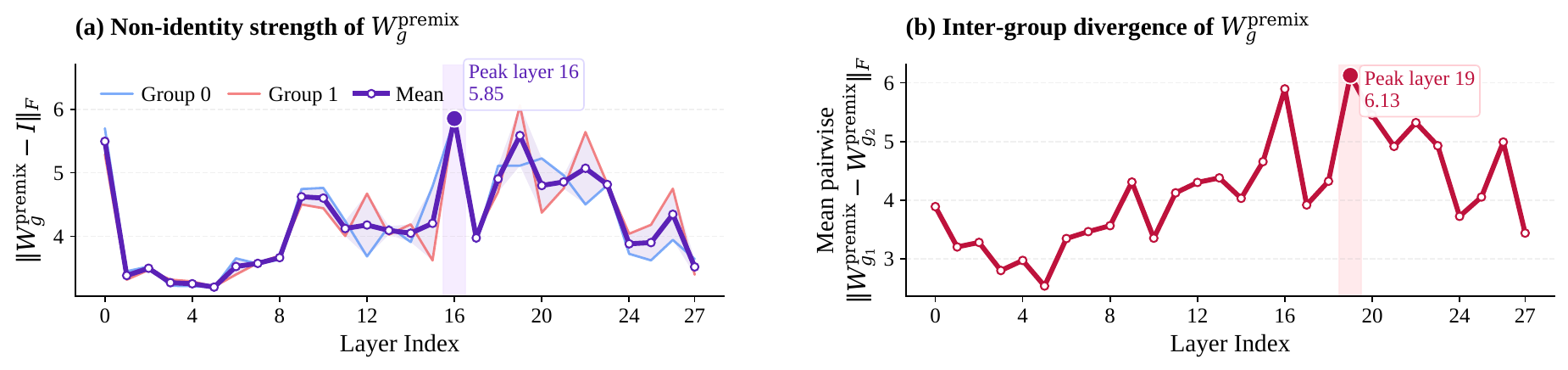}
\vskip -0.2cm
\caption{\textbf{Q Premix weight analysis on a trained Qwen-1.7B VHA model.}
(a)~Per-layer Frobenius deviation $\|W_g^{\mathrm{premix}} - I\|_F$ per KV group (group mean overlaid).
(b)~Per-layer mean pairwise distance $\|W_{g_1}^{\mathrm{premix}} - W_{g_2}^{\mathrm{premix}}\|_F$ between KV groups.}
\label{fig:premix}
\vskip -0.4cm
\end{figure}

\paragraph{Premix learns non-trivial, group-differentiated query transforms.}
Premix introduces a learnable per-group query transform $W_g^{\mathrm{premix}}$ before RoPE, initialized near identity. Two trivial outcomes would falsify our hypothesis: (a)~$W_g^{\mathrm{premix}}$ remains near identity after training, rendering the module a no-op; or (b)~all KV groups converge to the same matrix, degenerating Premix into a group-agnostic shared adapter without group-conditioned specialization. Figure~\ref{fig:premix} refutes both: panel~(a) shows that $\| W_g^{\mathrm{premix}} - I \|_F$ is substantially nonzero and most pronounced in mid-to-late layers (peak 5.85), and panel~(b) shows that the inter-group distance $\| W_{g_1}^{\mathrm{premix}} - W_{g_2}^{\mathrm{premix}} \|_F$ is of comparable magnitude (peak 6.13), yielding a ratio well below the $\sqrt{2}$ value expected under fully independent groups and far above the zero expected under a shared deviation. This quantifies that the two KV groups learn substantially but not independently different query transforms. Premix therefore learns differentiated query probing directions per KV group under a fixed minimal KV cache, realizing genuine group-conditioned virtual-head expansion rather than mere parameter augmentation.

\begin{figure}[t]
\centering
\includegraphics[width=\linewidth]{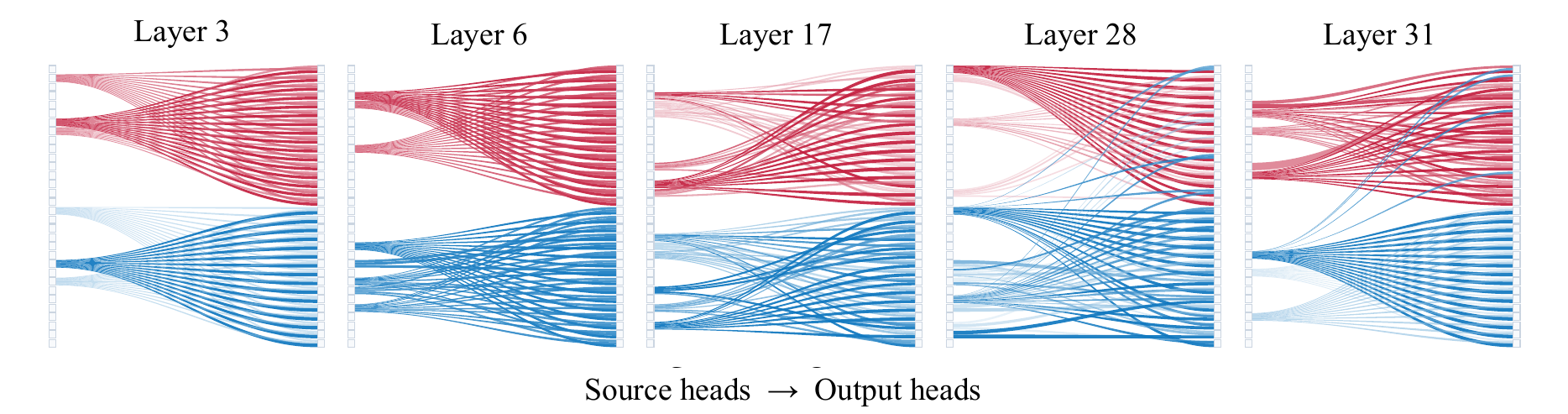}
\vskip -0.4cm
\caption{\textbf{Postmix head-mixing analysis on a trained Qwen-4B VHA model.}
Sankey diagrams of $|AB^\top|$ across five representative layers, with ribbons flowing from source heads (left, colored by KV group) to recombined output heads (right); width and color saturation proportional to $|AB^\top|_{j,i}$.}
\label{fig:postmix}
\vskip -0.6cm
\end{figure}

\paragraph{Postmix learns structured low-rank cross-head communication.}
Postmix recombines the $H_v$ attention output heads via the low-rank residual $I + BA^\top$, where entry $(j, i)$ of $|AB^\top|$ encodes the contribution of source head $j$ to recombined head $i$. The central question is whether $AB^\top$ merely performs a per-head rescaling (a strictly diagonal pattern) or genuinely redistributes information across heads (off-diagonal flow). Figure~\ref{fig:postmix} visualizes $|AB^\top|$ for the trained 4B model and reveals pronounced off-diagonal flow, ruling out the trivial per-head-rescaling outcome. The dominant pattern is \emph{within-group} redistribution: Postmix primarily rebalances information among heads sharing the same KV cache, with a sparse but consistent set of cross-group bridges providing inter-group connectivity without additional KV heads.

\paragraph{Joint reparameterization closes the doubly-compressed gap.}
Tying these analyses back to the component ablation in Table~\ref{tab:ablation} clarifies the three numbers reported there. Premix-only retains the group-conditioned query geometry of Figure~\ref{fig:premix} but lacks Postmix's cross-group repair pathway, hence the $-$0.78\% gap at 1.7B. Postmix-only owns the cross-head mixing channels of Figure~\ref{fig:postmix} but operates over collapsed, unspecialized virtual heads, hence the $-$0.14\% gap. Only the joint configuration realizes both structures simultaneously, yielding a $+$0.41\% advantage over GQA-8 at $25\%$ of its KV cache. VHA's effectiveness therefore arises from the functional complementarity of its two modules: Premix and Postmix are not interchangeable, and neither suffices alone.

\section{Conclusion}

We have presented Virtual Head Attention (VHA), which mitigates the tension among KV cache size, training/prefill FLOPs, and model quality through two lightweight linear operations: Q Premix and Linear Postmix. By halving physical query heads and recovering expressiveness via virtual head expansion and low-rank cross-group mixing, VHA reduces the KV cache to its architectural minimum while simultaneously lowering training FLOPs and improving model quality. Since all added operations are strictly linear, VHA fully absorbs into standard GQA-2 at inference, reusing existing serving optimizations with zero additional overhead. Analysis of trained models reveals that Premix and Postmix are functionally complementary and neither suffices alone, explaining why the joint configuration consistently surpasses the GQA baseline across all five evaluated scales (0.6B to 30B-A3B MoE). A matched 10B-A1B evaluation further shows that VHA maintains this advantage over GQA through 128K context.

\section*{Acknowledgments}

We thank Wenbin Chang for valuable input and helpful discussions. We also thank the anonymous reviewers, area chairs, and senior area chairs for their insightful comments and constructive suggestions.

This work was supported by the National Engineering Research Center of Deep Learning Technology and Application and the Fundamental and Interdisciplinary Disciplines Breakthrough Plan of the Ministry of Education of China (Grant No. JYB2025XDXM202).

\clearpage
\bibliographystyle{plainnat}
\bibliography{references}

\clearpage
\section*{Limitations and Future Work}

Our largest validation is at the 30B-A3B MoE scale trained on 500B tokens; scaling VHA to larger MoE models and substantially more training data (e.g., 20T tokens) remains an important direction. We further validate VHA through 128K on a 10B-A1B MoE, while broader long-context architectures and training recipes remain to be explored. VHA's behavior under reinforcement learning training paradigms is also worth exploring. Finally, VHA is orthogonal to other KV cache compression techniques such as quantization and sparsification, and combining them may yield further inference savings.

\FloatBarrier
\appendix

\section{Long-Context Evaluation}
\label{app:long-context}

\subsection{Model and Stage-wise Training}

We conduct a controlled long-context comparison on a MiMo-2.5-Flash-style 10B-A1B MoE backbone that interleaves full and sliding-window attention at a 1:3 ratio. The GQA and VHA models share the same backbone, data, optimization settings, and sliding-window layers. They differ only in the full-attention layers: GQA uses four KV heads, whereas VHA uses two; all sliding-window layers use eight KV heads in both models. Thus, VHA halves the KV cache of the full-attention layers without changing the sliding-window component.

The 8K pretraining stage uses 500B tokens and ends at 60K steps. Long-context mid-training then uses 100B tokens at 32K followed by another 100B tokens at 128K. These stages add 12K and 24K steps, yielding cumulative checkpoints at 72K and 96K steps, respectively. The RoPE base is increased from $10^4$ at 8K to $6.4{\times}10^5$ at 32K and $5{\times}10^6$ at 128K. We report intermediate results from the 32K checkpoint and final results from the 128K checkpoint.

\subsection{RULER and HELMET}

We evaluate RULER and HELMET from 8K to 128K. RULER Overall is the mean over 13 tasks: eight Retrieval, two Aggregation, two QA, and one Variable Tracking task. HELMET is reported by category because its categories measure distinct capabilities.

\begin{table*}[t]
\centering
\scriptsize
\caption{Long-context results on RULER and HELMET (\%). RULER Overall averages all 13 tasks. Bold marks the better result within each matched checkpoint and evaluation length; exact ties are unbolded.}
\label{tab:long-context-ruler-helmet}
\setlength{\tabcolsep}{3pt}
\renewcommand{\arraystretch}{1.04}
\resizebox{\textwidth}{!}{%
\begin{tabular}{lllccccc ccccc}
\toprule
& & & \multicolumn{5}{c}{RULER} & \multicolumn{5}{c}{HELMET} \\
\cmidrule(lr){4-8}\cmidrule(lr){9-13}
Checkpoint & Length & Model & Overall & Retr. & Aggr. & QA & VT & RAG & LongQA & Summ. & ICL & Rerank \\
\midrule
32K
& 8K & GQA & 80.9 & 92.4 & 73.1 & 56.5 & 53.6 & 24.5 & 9.0 & \textbf{7.2} & \textbf{60.7} & 0.3 \\
&    & VHA & \textbf{85.7} & \textbf{95.9} & \textbf{78.0} & \textbf{60.1} & \textbf{69.8} & \textbf{29.6} & \textbf{11.0} & 6.6 & 57.1 & \textbf{0.4} \\
& 16K & GQA & 73.7 & 90.0 & 51.8 & 53.7 & 27.6 & 22.6 & \textbf{11.0} & \textbf{8.0} & \textbf{70.4} & 0.0 \\
&     & VHA & \textbf{83.2} & \textbf{93.3} & \textbf{71.5} & \textbf{55.4} & \textbf{80.8} & \textbf{32.3} & 10.5 & 7.2 & 68.6 & 0.0 \\
& 32K & GQA & 69.7 & 81.9 & 51.9 & 49.0 & 49.4 & 22.3 & \textbf{17.0} & \textbf{8.1} & \textbf{78.3} & 0.0 \\
&     & VHA & \textbf{73.5} & \textbf{83.5} & \textbf{56.8} & \textbf{51.8} & \textbf{71.4} & \textbf{33.4} & 11.0 & 7.6 & 75.1 & \textbf{0.5} \\
\midrule
128K
& 8K & GQA & 79.1 & 89.5 & 87.9 & 46.0 & 44.6 & 27.4 & 12.5 & \textbf{5.9} & \textbf{58.7} & \textbf{0.6} \\
&    & VHA & \textbf{90.1} & \textbf{96.7} & \textbf{92.8} & \textbf{59.4} & \textbf{94.0} & \textbf{34.2} & 12.5 & 5.6 & 46.1 & 0.1 \\
& 16K & GQA & 78.2 & 88.1 & 95.1 & 46.7 & 28.4 & 28.4 & 13.0 & \textbf{6.4} & \textbf{66.4} & 0.5 \\
&     & VHA & \textbf{88.5} & \textbf{93.5} & \textbf{98.9} & \textbf{55.9} & \textbf{93.4} & \textbf{34.3} & \textbf{14.5} & 6.3 & 61.0 & \textbf{0.7} \\
& 32K & GQA & 72.6 & 83.3 & 93.6 & 43.2 & 3.2 & 26.7 & \textbf{21.0} & 6.7 & \textbf{75.6} & 0.4 \\
&     & VHA & \textbf{85.3} & \textbf{89.6} & \textbf{96.0} & \textbf{53.7} & \textbf{92.6} & \textbf{36.9} & 18.5 & \textbf{6.9} & 72.1 & 0.4 \\
& 64K & GQA & 69.5 & 81.0 & 86.4 & 39.5 & 4.2 & 27.8 & 19.0 & \textbf{7.9} & \textbf{80.1} & \textbf{0.7} \\
&     & VHA & \textbf{81.3} & \textbf{85.6} & \textbf{90.2} & \textbf{52.2} & \textbf{87.2} & \textbf{39.5} & \textbf{20.0} & 7.7 & 79.6 & 0.4 \\
& 128K & GQA & 65.7 & 75.3 & 92.5 & 31.5 & 3.0 & 30.4 & 13.0 & 7.8 & \textbf{82.6} & 2.0 \\
&      & VHA & \textbf{76.8} & \textbf{79.9} & 92.5 & \textbf{45.5} & \textbf{83.0} & \textbf{42.5} & \textbf{15.0} & \textbf{9.5} & 82.2 & \textbf{2.6} \\
\bottomrule
\end{tabular}%
}
\end{table*}

At the final checkpoint, VHA leads RULER Overall at every evaluated length, with an average of 84.4 across lengths versus 73.0 for GQA. The largest gap occurs in Variable Tracking, where VHA scores 92.6/87.2/83.0 at 32K/64K/128K versus 3.2/4.2/3.0 for GQA. HELMET is more heterogeneous: averaged across lengths, VHA wins RAG, LongQA, and Summarization, ties Reranking, and trails on ICL.

\subsection{Complementary Long-Context Evaluations}

We additionally report LongBench-v2, InfiniteBench, BABILong, a short-context eight-task average, and language-modeling bits per byte on the four ProLong domains. At the final checkpoint, VHA wins seven of the eight reported metrics; InfiniteBench is nearly tied.

\begin{table*}[t]
\centering
\footnotesize
\caption{Complementary evaluation results. LB-v2, InfBench, and BABI denote LongBench-v2, InfiniteBench, and BABILong; Short is the short-context eight-task average. The remaining columns report bits per byte (BPB) on ProLong domains. Bold marks the better matched result.}
\label{tab:long-context-complementary}
\setlength{\tabcolsep}{5pt}
\renewcommand{\arraystretch}{1.05}
\resizebox{\textwidth}{!}{%
\begin{tabular}{llcccccccc}
\toprule
Checkpoint & Model & LB-v2 $\uparrow$ & InfBench $\uparrow$ & BABI $\uparrow$ & Short $\uparrow$ & arXiv $\downarrow$ & Books $\downarrow$ & Code $\downarrow$ & Web $\downarrow$ \\
\midrule
32K & GQA & \textbf{28.0} & \textbf{29.9} & 36.5 & 62.7 & 0.568 & 0.812 & \textbf{0.294} & \textbf{0.763} \\
    & VHA & 22.3 & 27.3 & \textbf{37.4} & \textbf{63.7} & 0.568 & 0.812 & 0.295 & 0.765 \\
\midrule
128K & GQA & 18.9 & \textbf{31.8} & 31.1 & 62.6 & 0.582 & 0.819 & 0.311 & 0.768 \\
     & VHA & \textbf{24.2} & 31.5 & \textbf{38.2} & \textbf{64.1} & \textbf{0.564} & \textbf{0.804} & \textbf{0.296} & \textbf{0.760} \\
\bottomrule
\end{tabular}%
}
\end{table*}

\section{Additional Robustness Experiments}
\label{app:robustness}

The following experiments use a more efficient training implementation than the main 50B-token runs. Absolute scores therefore need not match the main tables, but all methods within each comparison use the same implementation and training recipe.

\subsection{Repeated-run Robustness at 0.6B}

We repeat the 0.6B experiment four times for VHA and three times for GQA. We report every run using zero-shot BF16 evaluation, with normalized accuracy used where available. Avg is the unweighted mean over eight tasks, and $\pm$ denotes the sample standard deviation across run-level averages. VHA achieves a higher mean with lower observed dispersion.

\begin{table*}[t]
\centering
\scriptsize
\caption{Repeated-run results at 0.6B (\%). Every run is reported. Bold in the summary rows marks the higher across-run mean.}
\label{tab:repeated-runs}
\setlength{\tabcolsep}{3.5pt}
\renewcommand{\arraystretch}{1.04}
\resizebox{\textwidth}{!}{%
\begin{tabular}{llccccccccc}
\toprule
Model & Run & ARC-C & BoolQ & HSwag & OBQA & PIQA & Wino & SciQ & SIQA & Avg \\
\midrule
VHA & 01 & 30.12 & 60.18 & 44.46 & 33.80 & 68.77 & 52.25 & 74.20 & 39.56 & 50.42 \\
VHA & 02 & 29.35 & 59.27 & 44.36 & 32.80 & 68.55 & 55.96 & 74.00 & 39.56 & 50.48 \\
VHA & 03 & 29.27 & 59.36 & 44.79 & 33.80 & 67.14 & 52.25 & 76.00 & 40.23 & 50.35 \\
VHA & 04 & 30.12 & 56.09 & 44.68 & 34.80 & 68.55 & 52.64 & 76.10 & 39.76 & 50.34 \\
\multicolumn{2}{l}{\textbf{VHA mean}} & 29.72 & 58.73 & \textbf{44.57} & \textbf{33.80} & 68.25 & \textbf{53.28} & \textbf{75.08} & \textbf{39.78} & \textbf{$50.40{\pm}0.07$} \\
\midrule
GQA & 01 & 29.44 & 58.59 & 44.49 & 32.40 & 68.28 & 51.14 & 75.50 & 38.79 & 49.83 \\
GQA & 02 & 30.12 & 57.86 & 44.55 & 32.60 & 68.34 & 50.91 & 74.10 & 39.87 & 49.79 \\
GQA & 03 & 30.20 & 60.55 & 44.61 & 35.00 & 68.99 & 51.93 & 72.50 & 39.30 & 50.39 \\
\multicolumn{2}{l}{\textbf{GQA mean}} & \textbf{29.92} & \textbf{59.00} & 44.55 & 33.33 & \textbf{68.54} & 51.33 & 74.03 & 39.32 & $50.00{\pm}0.34$ \\
\bottomrule
\end{tabular}%
}
\end{table*}

\subsection{Sensitivity to Premix Initialization}

We train the 1.7B model from scratch with the same recipe while changing only the Premix initialization: the default near-identity initialization ($\alpha{=}0.1$), a larger identity-anchored perturbation ($\alpha{=}1.0$), and fully random initialization without the identity anchor. All three runs remain stable, with no gradient explosion, divergence, NaN, or Inf. Their early loss and pre-clipping gradient norms are similar (Table~\ref{tab:init-trajectory}), while the default near-identity initialization gives the highest downstream average (Table~\ref{tab:init-downstream}).

\begin{table}[h]
\centering
\scriptsize
\caption{Early training trajectory under different Premix initializations. Each entry is loss / pre-clipping gradient norm.}
\label{tab:init-trajectory}
\setlength{\tabcolsep}{5pt}
\renewcommand{\arraystretch}{1.03}
\begin{tabular}{rccc}
\toprule
Step & $\alpha{=}0.1$ & $\alpha{=}1.0$ & Random \\
\midrule
10   & 12.2976 / 14.74 & 12.3072 / 14.80 & 12.3132 / 14.98 \\
20   & 11.1501 / 15.97 & 11.1581 / 16.26 & 11.1576 / 16.25 \\
30   & 9.9346 / 4.98   & 9.9389 / 5.01   & 9.9400 / 5.07 \\
50   & 9.1168 / 2.67   & 9.1243 / 3.76   & 9.1255 / 2.03 \\
100  & 7.9670 / 2.96   & 7.9788 / 2.93   & 7.9669 / 4.02 \\
200  & 6.4777 / 2.99   & 6.4878 / 2.21   & 6.4898 / 1.65 \\
500  & 4.9822 / 1.70   & 4.9990 / 2.47   & 5.0198 / 2.85 \\
1000 & 3.8647 / 1.17   & 3.8851 / 0.94   & 3.8888 / 0.89 \\
2000 & 3.1699 / 0.27   & 3.1842 / 0.33   & 3.1967 / 0.54 \\
\bottomrule
\end{tabular}
\end{table}

\begin{table}[h]
\centering
\scriptsize
\caption{Downstream accuracy under different Premix initializations (\%). Bold marks the best result per row.}
\label{tab:init-downstream}
\setlength{\tabcolsep}{7pt}
\renewcommand{\arraystretch}{1.03}
\begin{tabular}{lccc}
\toprule
Task & $\alpha{=}0.1$ & $\alpha{=}1.0$ & Random \\
\midrule
ARC-C      & \textbf{34.81} & 34.13 & 34.39 \\
BoolQ      & \textbf{60.06} & 58.04 & 57.03 \\
HellaSwag  & \textbf{51.80} & 51.72 & 51.76 \\
OpenBookQA & \textbf{37.60} & 37.00 & 36.00 \\
PIQA       & 71.06 & 71.33 & \textbf{72.14} \\
Winogrande & \textbf{55.72} & 54.85 & 53.28 \\
SciQ       & \textbf{80.70} & 80.30 & 80.30 \\
Social IQA & \textbf{40.58} & 40.23 & 39.97 \\
\midrule
Avg        & \textbf{54.04} & 53.45 & 53.11 \\
\bottomrule
\end{tabular}
\end{table}

\section{Per-benchmark Scores for the 30B-A3B MoE Evaluation}
\label{app:per-benchmark}

Table~\ref{tab:per-benchmark-moe} reports per-benchmark scores for all evaluated models on the 30B-A3B MoE suite.
Category-averaged numbers are reproduced in Table~\ref{tab:large-scale} in the main text.
Model configurations: GQA 48-4-4, MLA 28-28-28, VHA 24-2-2.
\textbf{Bold} marks the best per row.

\begin{table*}[h]
\centering
\footnotesize
\caption{Per-benchmark scores on the 30B-A3B MoE evaluation suite (accuracy \%).}
\label{tab:per-benchmark-moe}
\setlength{\tabcolsep}{5pt}
\renewcommand{\arraystretch}{1.08}
\begin{tabular}{llcrrr}
\toprule
Category & Benchmark & Shot & GQA & MLA & \textbf{VHA} \\
\midrule
\multirow{4}{*}{General}
 & MMLU        & 5-shot        & 55.97          & 59.34          & \textbf{60.24} \\
 & MMLU-Redux  & 5-shot        & 63.05          & 65.25          & \textbf{66.10} \\
 & MMLU-Pro    & 5-shot        & 36.20          & \textbf{41.12} & 40.03          \\
 & BBH         & 3-shot CoT    & 56.23          & 56.57          & \textbf{58.61} \\
\midrule
\multirow{3}{*}{Math \& STEM}
 & GPQA        & 10-shot CoT   & \textbf{34.62} & 31.62          & 34.11          \\
 & GSM8K       & 4-shot CoT    & \textbf{56.48} & 55.42          & 56.25          \\
 & MATH        & 4-shot CoT    & 38.18          & 43.42          & \textbf{48.84} \\
\midrule
\multirow{3}{*}{Coding}
 & HumanEval+  & 0-shot        & 50.00          & \textbf{52.44} & 47.56          \\
 & MBPP        & 3-shot        & 45.00          & 48.60          & \textbf{49.20} \\
 & MBPP+       & 3-shot        & 60.54          & 62.07          & \textbf{62.93} \\
\midrule
\multirow{2}{*}{Multilingual}
 & MMMLU       & 5-shot        & 39.82          & 42.69          & \textbf{42.95} \\
 & INCLUDE     & 5-shot        & 46.69          & 47.71          & \textbf{48.16} \\
\bottomrule
\end{tabular}
\end{table*}

\section{Training Configuration and Benchmark Citations}
\label{app:setup}

\subsection{Model Architecture and Training Hyperparameters}
\label{app:model-train-config}

Table~\ref{tab:full-config} lists the backbone architecture and training
hyperparameters for all five scales.
Small-scale models follow the Qwen3 architecture; the 30B-A3B model uses
an ERNIE-style MoE architecture with auxiliary-free routing
(\texttt{noaux\_tc}) and z-loss ($\lambda{=}4{\times}10^{-4}$).
We use AdamW \citep{loshchilov2017decoupled} with $\beta_1{=}0.9$,
$\beta_2{=}0.95$, weight decay $0.1$, and gradient clipping at $1.0$.

\begin{table*}[h]
\centering
\footnotesize
\caption{Model architecture and training hyperparameters across all five scales.
``--'' denotes not applicable.}
\label{tab:full-config}
\setlength{\tabcolsep}{5pt}
\renewcommand{\arraystretch}{1.05}
\begin{tabular}{l | ccccc}
\toprule
 & \textbf{0.6B} & \textbf{1.7B} & \textbf{4B} & \textbf{8B-A1B (MoE)} & \textbf{30B-A3B (MoE)} \\
\specialrule{0em}{2pt}{1pt}
\midrule
\specialrule{0em}{2pt}{1pt}
\multicolumn{6}{l}{\textit{Architecture}} \\
\specialrule{0em}{1pt}{1pt}
Precision               & BF16   & BF16   & BF16   & BF16          & FP8            \\
\specialrule{0em}{0.5pt}{1pt}
Num layers              & 28     & 28     & 36     & 24            & 46             \\
\specialrule{0em}{0.5pt}{1pt}
Hidden size             & 1024   & 2048   & 2560   & 2048          & 1536           \\
\specialrule{0em}{0.5pt}{1pt}
FFN dim                 & 3072   & 6144   & 9728   & $768{\times}64$ & $1024{\times}128$ \\
\specialrule{0em}{0.5pt}{1pt}
Num experts             & --     & --     & --     & 64            & 128 (+1 shared)  \\
\specialrule{0em}{0.5pt}{1pt}
Experts per token       & --     & --     & --     & 4             & 8              \\
\specialrule{0em}{0.5pt}{1pt}
Vocab size              & 151936 & 151936 & 151936 & 151936        & 102400         \\
\specialrule{0em}{0.5pt}{1pt}
RoPE base               & $10^6$ & $10^6$ & $10^6$ & $10^6$        & $10^6$         \\
\specialrule{0em}{0.5pt}{1pt}
Tied embeddings         & \checkmark & \checkmark & \checkmark & $\times$ & $\times$ \\
\specialrule{0em}{2pt}{1pt}
\midrule
\specialrule{0em}{2pt}{1pt}
\multicolumn{6}{l}{\textit{Training}} \\
\specialrule{0em}{1pt}{1pt}
Training tokens         & 50B    & 50B    & 50B    & 50B           & 500B           \\
\specialrule{0em}{0.5pt}{1pt}
Context length          & 4096   & 4096   & 4096   & 4096          & 8192 \\
\specialrule{0em}{0.5pt}{1pt}
Peak learning rate      & $2.5{\times}10^{-4}$ & $2{\times}10^{-4}$ & $2{\times}10^{-4}$ & $2{\times}10^{-4}$ & $1{\times}10^{-4}$ \\
\specialrule{0em}{0.5pt}{1pt}
Min learning rate       & $1{\times}10^{-5}$ & $1{\times}10^{-5}$ & $1{\times}10^{-5}$ & $1{\times}10^{-5}$ & $1{\times}10^{-5}$   \\
\specialrule{0em}{0.5pt}{1pt}
LR schedule             & cosine & cosine & cosine & cosine         & wsd   \\
\specialrule{0em}{0.5pt}{1pt}
Warmup steps            & 2000   & 2000   & 2000   & 2000           & 2000   \\
\specialrule{0em}{0.5pt}{1pt}
Total steps             & 24000  & 24000  & 24000  & 24000          & 40000   \\
\specialrule{0em}{0.5pt}{1pt}
Batch size (tokens)     & 2M     & 2M     & 2M     & 2M             & 13.1M   \\
\specialrule{0em}{0.5pt}{1pt}
Adam $\beta_1$          & 0.9    & 0.9    & 0.9    & 0.9           & 0.9            \\
\specialrule{0em}{0.5pt}{1pt}
Adam $\beta_2$          & 0.95   & 0.95   & 0.95   & 0.95          & 0.95           \\
\specialrule{0em}{0.5pt}{1pt}
Weight decay            & 0.1    & 0.1    & 0.1    & 0.1           & 0.1            \\
\specialrule{0em}{0.5pt}{1pt}
Gradient clipping       & 1.0    & 1.0    & 1.0    & 1.0           & 1.0            \\
\specialrule{0em}{2pt}{1pt}
\midrule
\specialrule{0em}{2pt}{1pt}
\multicolumn{6}{l}{\textit{VHA-specific}} \\
\specialrule{0em}{1pt}{1pt}
Premix init scale $\alpha$  & 0.1 & 0.1 & 0.1 & 0.1 & 0.1 \\
\specialrule{0em}{0.5pt}{1pt}
Premix init noise $\sigma$  & 1 & 1 & 1 & 1 & 1 \\
\specialrule{0em}{0.5pt}{1pt}
Postmix rank $r$            & 4 & 4 & 8 & 8 & 12 \\
\specialrule{0em}{2pt}{1pt}
\midrule
\specialrule{0em}{2pt}{1pt}
\multicolumn{6}{l}{\textit{Baseline-specific}} \\
\specialrule{0em}{1pt}{1pt}
MLA latent dim $c_{kv}$     & 512 & 512 & 512 & 512 & 512 \\
\specialrule{0em}{0.5pt}{1pt}
MLA latent dim $c_{q}$      & 512 & 1536 & 1536 & 1536 & 1536 \\
\specialrule{0em}{0.5pt}{1pt}
MFA head dim $d'$           & 256 & 256 & 256 & 256 & -- \\
\specialrule{0em}{2pt}{1pt}
\bottomrule
\end{tabular}
\end{table*}

\subsection{Evaluation Benchmarks}

\paragraph{Small-scale benchmarks (8 tasks).}
We report zero-shot accuracy on the following benchmarks:
ARC-Challenge \citep{clark2018think},
BoolQ \citep{clark2019boolq},
HellaSwag \citep{zellers2019hellaswag},
OpenBookQA \citep{mihaylov2018can},
PIQA \citep{bisk2020piqa},
Winogrande \citep{sakaguchi2021winogrande},
SciQ \citep{welbl2017crowdsourcing},
Social IQA \citep{sap2019social}.

\paragraph{Large-scale benchmarks (12 tasks).}
We report accuracy with the shot counts listed in Table~\ref{tab:per-benchmark-moe} on the following benchmarks:
MMLU \citep{hendrycks2020measuring},
MMLU-Redux \citep{gema2025we},
MMLU-Pro \citep{wang2024mmlu},
BBH \citep{suzgun2023challenging},
GPQA \citep{rein2023gpqa},
GSM8K \citep{cobbe2021training},
MATH \citep{hendrycks2021measuring},
HumanEval+ \citep{liu2023your},
MBPP \citep{austin2021program},
MBPP+ \citep{liu2023your},
MMMLU \citep{openai2024mmmlu},
INCLUDE \citep{romanou2024include}.

\section{Training Loss Curves}
\label{app:loss-curves}

\begin{figure}[h]
  \centering
  \includegraphics[width=\linewidth]{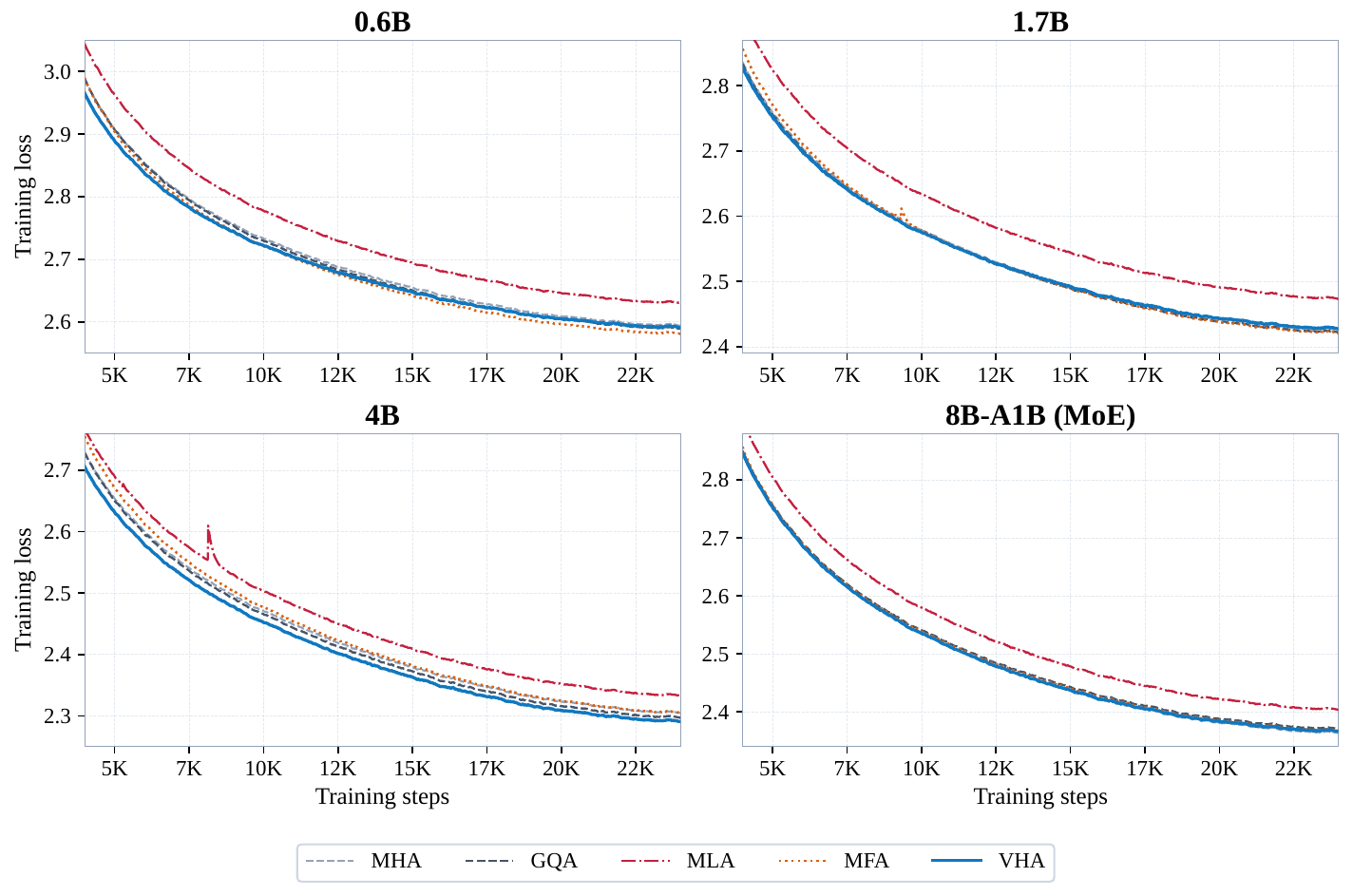}
  \caption{%
    \textbf{Training loss curves across four model scales.}
    Each panel compares MHA, GQA, MLA, MFA, and VHA on FineWeb-Edu
    (50B tokens, 4K context).
    All curves are plotted from step 4K onward after the learning-rate warm-up;
    smoothing is applied with an exponential moving average ($\alpha{=}0.93$).
    VHA (blue, solid) achieves the lowest loss at convergence at the 4B scale.%
  }
  \label{fig:loss-curves}
\end{figure}

Figure~\ref{fig:loss-curves} shows per-method training loss curves at four
model scales (50B tokens, 4K context).
VHA achieves the lowest convergence loss at the 4B scale and remains
competitive at 0.6B, 1.7B, and 8B-A1B, where its gap from the best-converging method is within 0.01 nats.

\section{VHA Algorithm}
\label{app:algorithm}

Algorithm~\ref{alg:vha} provides a self-contained reference for VHA's training
forward pass and the one-time inference-absorption procedure that reduces the
trained model to a standard GQA-2 deployment.

\begin{algorithm}[h]
\caption{Virtual Head Attention (VHA)}
\label{alg:vha}
\textbf{Training Forward Pass}
\begin{algorithmic}[1]
\REQUIRE Input $X \in \mathbb{R}^{S \times h}$; parameters $W_Q, W_K, W_V, W_O, \{W_g^{\mathrm{premix}}\}_{g=1}^{H_{kv}}, A, B$
\STATE $Q \leftarrow X W_Q \in \mathbb{R}^{S \times H_q' \times d}$ \hfill $\triangleright$ Physical query heads ($H_q' {=} H_q / 2$)
\STATE $K \leftarrow X W_K \in \mathbb{R}^{S \times H_{kv} \times d}$, \quad $V \leftarrow X W_V \in \mathbb{R}^{S \times H_{kv} \times d}$
\FOR{each KV group $g = 1, \dots, H_{kv}$}
    \STATE $\tilde{Q}_g \leftarrow Q \, W_g^{\mathrm{premix}}$ \hfill $\triangleright$ Q Premix: virtual head expansion
\ENDFOR
\STATE $\tilde{Q} \leftarrow \mathrm{concat}(\tilde{Q}_1, \dots, \tilde{Q}_{H_{kv}}) \in \mathbb{R}^{S \times H_v \times d}$ \hfill $\triangleright$ $H_v {=} H_{kv} {\times} H_q'$ virtual heads
\STATE $Y_g \leftarrow \mathrm{softmax}\!\left(\tilde{Q}_g K_g^\top / \sqrt{d}\right) V_g$ for each group $g$ \hfill $\triangleright$ Grouped attention
\STATE $Y \leftarrow \mathrm{concat}(Y_1, \dots, Y_{H_{kv}})$
    \STATE $Y' \leftarrow (I_{H_v} + B A^\top)\, Y$ \hfill $\triangleright$ Linear Postmix: cross-group mixing
\STATE $\mathrm{output} \leftarrow \mathrm{reshape}(Y', [S, H_v d]) \; W_O$
\end{algorithmic}

\vspace{0.4em}
\hrule
\vspace{0.4em}

\textbf{Inference Absorption (one-time precomputation)}
\begin{algorithmic}[1]
\REQUIRE $W_Q \in \mathbb{R}^{h \times H_q' d}$ (sliced into per-head $W_Q^{(i)} \in \mathbb{R}^{h \times d}$);\, $\{W_g^{\mathrm{premix}}\}_{g=1}^{H_{kv}}$ each $\in \mathbb{R}^{d \times d}$;\, $W_O \in \mathbb{R}^{H_v d \times h}$;\, $A, B \in \mathbb{R}^{H_v \times r}$
\FOR{each $(i, g) \in \{1, \dots, H_q'\} \times \{1, \dots, H_{kv}\}$}
    \STATE $\widetilde{W}_Q^{(i,g)} \leftarrow W_Q^{(i)} \, W_g^{\mathrm{premix}} \;\in\; \mathbb{R}^{h \times d}$ \hfill $\triangleright$ Fold Premix into the $(i,g)$-th virtual query projection
\ENDFOR
\STATE $W_Q^{\mathrm{abs}} \leftarrow \mathrm{concat}\!\left(\big\{\widetilde{W}_Q^{(i,g)}\big\}_{i,g}\right) \;\in\; \mathbb{R}^{h \times H_v d}$ \hfill $\triangleright$ $H_v {=} H_{kv} {\times} H_q'$
\STATE $W_O^{\mathrm{abs}} \leftarrow \bigl((I_{H_v} + AB^\top) \otimes I_d\bigr)\, W_O \;\in\; \mathbb{R}^{H_v d \times h}$ \hfill $\triangleright$ Fold Postmix into $W_O$
\STATE Discard $W_Q, \{W_g^{\mathrm{premix}}\}, A, B$; run standard GQA-2 with $W_Q^{\mathrm{abs}}, W_K, W_V, W_O^{\mathrm{abs}}$
\end{algorithmic}
\end{algorithm}

\section{Premix as Structured Factorization of Virtual-Head Query Projections}
\label{app:premix_proof}

We show that, in the forward pass, Q Premix is mathematically equivalent to a standard multi-head attention with $H_v = H_{kv} H_q'$ virtual query heads, whose per-virtual-head query projections are constrained to a structured low-dimensional manifold. The training-time benefit of the separated parameterization arises from the coupled gradient structure induced by this factorization.

\paragraph{Setup.} Let $W_Q^{\mathrm{phys}} \in \mathbb{R}^{h \times H_q' d}$ be the physical query projection (with $H_q'$ heads, each of dimension $d$), and let $W_g^{\mathrm{premix}} \in \mathbb{R}^{d \times d}$ be the Premix matrix for KV group $g \in \{1, \dots, H_{kv}\}$. For physical head $i \in \{1, \dots, H_q'\}$ within group $g$, the virtual query is
\begin{equation}
    \tilde{Q}_{i,g} \;=\; Q_i \, W_g^{\mathrm{premix}} \;=\; X \, W_Q^{(i)} \, W_g^{\mathrm{premix}},
\end{equation}
where $W_Q^{(i)} \in \mathbb{R}^{h \times d}$ is the $i$-th slice of $W_Q^{\mathrm{phys}}$. The attention output for virtual head $(i, g)$ is
\begin{equation}
    \mathrm{head}_{i,g} \;=\; \mathrm{softmax}\!\left(\frac{\tilde{Q}_{i,g} K_g^\top}{\sqrt{d}}\right) V_g.
\end{equation}

\paragraph{Forward equivalence to standard multi-head attention.} Define
\begin{equation}
    \widetilde{W}_Q^{(i,g)} \;\triangleq\; W_Q^{(i)} \, W_g^{\mathrm{premix}} \;\in\; \mathbb{R}^{h \times d}.
\end{equation}
By the associativity of matrix multiplication, $\tilde{Q}_{i,g} = X \widetilde{W}_Q^{(i,g)}$, and therefore
\begin{equation}
    \mathrm{head}_{i,g} \;=\; \mathrm{softmax}\!\left(\frac{X \widetilde{W}_Q^{(i,g)} K_g^\top}{\sqrt{d}}\right) V_g
    \label{eq:premix_forward_equiv}
\end{equation}
holds as a pointwise identity for every input $X$ and every parameter assignment. Hence VHA is, at the level of input-output behavior, a standard multi-head attention with $H_v$ query heads and $H_{kv}$ KV heads, in which the $i$-th virtual head of group $g$ uses the projection matrix $\widetilde{W}_Q^{(i,g)}$.

\paragraph{The factorization defines a low-dimensional parameter manifold.} The standard $H_v$-head Q projection has $H_v h d = H_{kv} H_q' h d$ free parameters. The Premix factorization uses
\begin{equation}
    H_q' h d \;+\; H_{kv} d^2
\end{equation}
parameters, parameterized by $(W_Q^{\mathrm{phys}}, \{W_g^{\mathrm{premix}}\}_{g=1}^{H_{kv}})$. The set of attainable $\{\widetilde{W}_Q^{(i,g)}\}$ tuples is therefore a structured submanifold of the unconstrained parameter space:
\begin{equation}
    \mathcal{M} \;=\; \left\{ \big\{ W_Q^{(i)} W_g^{\mathrm{premix}} \big\}_{i,g} \;:\; W_Q^{(i)} \in \mathbb{R}^{h \times d}, \; W_g^{\mathrm{premix}} \in \mathbb{R}^{d \times d} \right\}.
\end{equation}
At the 1.7B scale ($h{=}2048$, $d{=}128$, $H_q'{=}8$, $H_{kv}{=}2$, $H_v{=}16$), the factorized parameterization uses $\sim$2.13M parameters, compared to $\sim$4.19M for the unconstrained $H_v$-head Q projection, a $\sim$50\% reduction. The corresponding Q-projection FLOPs are reduced by a similar factor, as the additional cost $2 H_{kv} H_q' d^2$ of applying Premix is much smaller than the savings $2 h H_q' (H_{kv} - 1) d$ achieved by halving the physical head count. This is the principal source of VHA's reduction in training and prefill FLOPs.

\paragraph{Reparameterization benefit via gradient coupling.} Although Eq.~\ref{eq:premix_forward_equiv} shows that the forward pass is algebraically equivalent to training a single matrix $\widetilde{W}_Q^{(i,g)}$ per virtual head, the gradients under the factorized parameterization are coupled across groups and physical heads:
\begin{align}
    \frac{\partial \mathcal{L}}{\partial W_Q^{(i)}} &= \sum_{g=1}^{H_{kv}} X^\top \, \frac{\partial \mathcal{L}}{\partial \tilde{Q}_{i,g}} \, \big(W_g^{\mathrm{premix}}\big)^\top, \\
    \frac{\partial \mathcal{L}}{\partial W_g^{\mathrm{premix}}} &= \sum_{i=1}^{H_q'} \big(X W_Q^{(i)}\big)^\top \, \frac{\partial \mathcal{L}}{\partial \tilde{Q}_{i,g}}.
\end{align}
Each $W_Q^{(i)}$ aggregates contributions from all $H_{kv}$ groups, modulated by the corresponding Premix matrices, while each $W_g^{\mathrm{premix}}$ receives gradient signal through all $H_q'$ physical heads. SGD on the factorized parameters therefore traverses the manifold $\mathcal{M}$ along directions induced by this coupled gradient structure, which differs from the trajectories obtained by independently optimizing $H_v$ unconstrained virtual-head matrices. This is the same source of implicit regularization that underlies LoRA \citep{hu2022lora} and RepVGG \citep{ding2021repvgg}: a forward-equivalent factorization of a large weight tensor yields different optimization dynamics and, empirically, different (and often better) final solutions.

\paragraph{Inference-time absorption.} Since the forward equivalence in Eq.~\ref{eq:premix_forward_equiv} holds for any frozen parameter values, at inference we simply precompute $\widetilde{W}_Q^{(i,g)} = W_Q^{(i)} W_g^{\mathrm{premix}}$ for all $(i, g)$ pairs and discard $W_Q^{\mathrm{phys}}$ and $\{W_g^{\mathrm{premix}}\}$. The model then runs as a standard GQA-2 with $H_v$ independent query projections, $H_{kv}=2$ KV heads, and the same output projection $W_O$, fully compatible with existing GQA serving kernels.

\paragraph{Unified view with Postmix.} Premix and Postmix are two instantiations of the same head-axis reparameterization principle. Premix factorizes the per-virtual-head query projection on the Q side ($\widetilde{W}_Q^{(i,g)} = W_Q^{(i)} W_g^{\mathrm{premix}}$); Postmix factorizes the merged output projection on the O side ($W_O' = \bigl((I_{H_v} + AB^\top) \otimes I_d\bigr) W_O$, see Appendix~\ref{app:postmix_proof}). Both are forward-equivalent to a standard GQA in their merged form, both gain training-time benefits from coupled gradients of the separated parameters, and both fold back into a standard GQA-2 architecture at inference with zero overhead.

\section{Proof of Postmix Linear Redundancy}
\label{app:postmix_proof}

We prove that Linear Postmix can be mathematically absorbed into the output projection $W_O$, and discuss why the separated parameterization nonetheless improves training.

\paragraph{Setup.} For a single token, let $Y \in \mathbb{R}^{H_v \times d}$ be the concatenated attention output across all $H_v$ virtual heads. Postmix computes:
\begin{equation}
    Y' = (I_{H_v} + BA^\top) \, Y
\end{equation}
where $A, B \in \mathbb{R}^{H_v \times r}$. The final output is obtained by applying the output projection:
\begin{equation}
    \mathrm{output}^\top = \mathrm{vec}_\text{row}(Y')^\top W_O \in \mathbb{R}^{1 \times h}
\end{equation}
where $W_O \in \mathbb{R}^{H_v d \times h}$ and $\mathrm{vec}_\text{row}(\cdot)$ flattens a matrix in row-major order (i.e., $Y\mathrm{.reshape}(H_v d)$ in PyTorch, concatenating head outputs).

\paragraph{Absorption via Kronecker product.} Using the row-major identity: for $M \in \mathbb{R}^{H_v \times H_v}$ and $Y \in \mathbb{R}^{H_v \times d}$,
\begin{equation}
    \mathrm{vec}_\text{row}(MY)^\top = \mathrm{vec}_\text{row}(Y)^\top \bigl(M^\top \otimes I_d\bigr)
    \label{eq:rowmajor_kron}
\end{equation}
Setting $M = I_{H_v} + BA^\top$ (so $M^\top = I_{H_v} + AB^\top$) \citep{roger1994topics}:
\begin{align}
    \mathrm{vec}_\text{row}(Y')^\top &= \mathrm{vec}_\text{row}\!\left((I_{H_v} + BA^\top)\,Y\right)^\top \notag \\
    &= \mathrm{vec}_\text{row}(Y)^\top \bigl((I_{H_v} + AB^\top) \otimes I_d\bigr)
\end{align}

Substituting into the output computation:
\begin{align}
    \mathrm{output}^\top &= \mathrm{vec}_\text{row}(Y')^\top W_O \notag \\
    &= \mathrm{vec}_\text{row}(Y)^\top \bigl((I_{H_v} + AB^\top) \otimes I_d\bigr) W_O \notag \\
    &= \mathrm{vec}_\text{row}(Y)^\top W_O'
\end{align}
where $W_O' = \bigl((I_{H_v} + AB^\top) \otimes I_d\bigr) W_O \in \mathbb{R}^{H_v d \times h}$.

Since $I_{H_v} + BA^\top$ is generically invertible (it equals the identity at initialization and remains near-identity during training), $W_O'$ has the same number of free parameters as $W_O$. Therefore, for any choice of $W_O, A, B$, there exists a single matrix $W_O'$ that produces the identical output, confirming that Postmix is theoretically redundant.

\paragraph{Reparameterization benefit.} Despite this redundancy, the separated parameterization $\{W_O, A, B\}$ provides training advantages analogous to those observed in RepVGG \citep{ding2021repvgg}:

\begin{enumerate}
    \item \textbf{Factored gradient paths.} The gradients with respect to $A$, $B$, and $W_O$ follow different paths through the computation graph. This factorization provides a form of implicit regularization, as the effective weight update to $W_O'$ at each step is constrained by the low-rank structure of $AB^\top$.
    \item \textbf{Identity initialization.} With $B$ initialized to zero, $BA^\top = 0$ and $Y' = Y$ at the start of training. Postmix begins as a no-op and gradually learns useful cross-group mixing patterns, avoiding disruption to the early training dynamics.
    \item \textbf{Low-rank constraint.} The bottleneck rank $r \ll H_v$ limits the expressiveness of the cross-head mixing, acting as a structural regularizer. This prevents Postmix from overfitting, consistent with the ablation finding that larger ranks ($r{=}32, 128$) degrade evaluation performance despite lower training loss.
\end{enumerate}

\end{document}